\documentclass[conference, twocolumn]{IEEEtran}
\usepackage{listings}
\usepackage{graphicx}
\usepackage{subcaption}
\usepackage[table,xcdraw]{xcolor}
\usepackage{amsmath}
\usepackage{pifont}
\usepackage{amssymb}
\definecolor{mydarkblue}{rgb}{0,0.08,0.65}
\usepackage[colorlinks=true,
    linkcolor=mydarkblue,
    citecolor=mydarkblue,
    filecolor=mydarkblue,
    urlcolor=mydarkblue]{hyperref} 
\usepackage{cleveref}
\usepackage{soul}
\usepackage[shortlabels]{enumitem}
\usepackage{url}
\usepackage{color}
\usepackage{tikz}
\usepackage{balance}
\usepackage{multirow}
\usepackage{comment}
\usepackage{wrapfig,lipsum,booktabs}
\usepackage[frozencache,cachedir=.]{minted}
\newcommand\numberthis{\addtocounter{equation}{1}\tag{\theequation}}
\usepackage[symbol]{footmisc}
\usepackage{tablefootnote}
\usepackage{tabularx}
\usepackage{placeins}
\usepackage{algorithm}
\usepackage{algpseudocode}

\usepackage{cuted}
\usepackage{float}
\usepackage{caption}

\usepackage{multicol}

\usepackage{dblfloatfix}
\usepackage{natbib}

\definecolor{codegreen}{rgb}{0,0.6,0}
\definecolor{codegray}{rgb}{0.5,0.5,0.5}
\definecolor{codepurple}{rgb}{0.58,0,0.82}
\definecolor{backcolour}{rgb}{0.95,0.95,0.92}

\makeatletter
\def\blfootnote{\xdef\@thefnmark{}\@footnotetext}
\makeatother

\lstdefinestyle{mystyle}{
  backgroundcolor=\color{backcolour},   commentstyle=\color{codegreen},
  keywordstyle=\color{magenta},
  numberstyle=\tiny\color{codegray},
  stringstyle=\color{codepurple},
  basicstyle=\ttfamily\footnotesize,
  breakatwhitespace=false,         
  breaklines=true,                 
  captionpos=b,                    
  keepspaces=true,                 
  numbers=left,                    
  numbersep=5pt,                  
  showspaces=false,                
  showstringspaces=false,
  showtabs=false,                  
  tabsize=2,
}

\AtBeginDocument{%
  \providecommand\BibTeX{{%
    \normalfont B\kern-0.5em{\scshape i\kern-0.25em b}\kern-0.8em\TeX}}}

\renewcommand{\IEEEauthorrefmark}[1]{\textsuperscript{#1}}

\IEEEoverridecommandlockouts
\begin{document}

\title{Generalizing E-prop to Deep Networks}

\newcommand{\corr}{\textsuperscript{*}}

\author{
\IEEEauthorblockN{Beren Millidge\IEEEauthorrefmark{1}}
\IEEEauthorblockA{\IEEEauthorrefmark{1}Zyphra}
\IEEEauthorblockA{\texttt{beren@zyphra.com}}
}

\maketitle

\setcounter{page}{1}

\begin{abstract}

Recurrent networks are typically trained with backpropagation through time (BPTT). However, BPTT requires storing the history of all states in the network and then replaying them sequentially backwards in time. This computation appears extremely implausible for the brain to implement. Real Time Recurrent Learning (RTRL) proposes an mathematically equivalent alternative where gradient information is propagated forwards in time locally alongside the regular forward pass, however it has significantly greater computational complexity than BPTT which renders it impractical for large networks. E-prop proposes an approximation of RTRL which reduces its complexity to the level of BPTT while maintaining a purely online forward update which can be implemented by an eligibility trace at each synapse.
However, works on RTRL and E-prop ubiquitously investigate learning in a single layer with recurrent dynamics. However, learning in the brain spans multiple layers and consists of both hierarchal dynamics in depth as well as time. In this mathematical note, we extend the E-prop framework to handle arbitrarily deep networks, deriving a novel recursion relationship across depth which extends the eligibility traces of E-prop to deeper layers. Our results thus demonstrate an online learning algorithm can perform accurate credit assignment across both time and depth simultaneously, allowing the training of deep recurrent networks without backpropagation through time. 
\end{abstract}

The cortical learning algorithm in the brain operates over both time and depth. Neural activity as a whole evolves with clear recurrent dynamics very different from the classical feedforward networks used in much of machine learning. However, how exactly the brain solves the credit assignment and learning problem across both time (recurrent dynamics) and depth (cortical layers and regions) is still unknown. 

While there exists a small literature addressing the question of biologically plausible learning algorithms in deep networks \citep{lillicrap2016random,scellier2017equilibrium,millidge2021predictive,lee2015difference,alonso2022theoretical,millidge2022backpropagation,meulemans2022least}, almost all of this literature focuses on the purely feedforward case. However, how precisely the brain can achieve accurate and efficient credit assignment across time is even more mysterious. The only widely-used and effective algorithm we have that can achieve this today is backpropagation through time (BPTT), which is known both to be fundamentally limited as well as wildly implausible for implementation in the brain \citep{lillicrap2019backpropagation}. BPTT ultimately unrolls the temporal dimension of the network and then treats it as if it were depth, requiring storing all the activations and gradients at every point in time and then systematically backpropagating backwards in time from the final answer to the initial conditions of the network. Such an approach is challenging and limiting even on digital computers since ultimately the memory and compute required for learning grow linearly with the sequence length ultimately rendering learning on long enough sequences impractical. However, it is entirely unrealistic for the brain to store it entire history of states and then replay them backwards in time to compute updates, nor is there any hint of real synaptic plasticity in-vivo actually acting in such a manner. Instead, the brain appears to use various mechanisms such as eligibility traces and growth and pruning of synapses and dendrites to perform credit assignment over time, with plasticity happening immediately or very soon after stimuli vs in distinct forward and backwards phases.

However, BPTT is not the only approach to credit assignment through time. For a long time, it has been recognized that if we instead use the analog of forward-mode differentiation, we can derive a learning algorithm that instead performs credit assignment \emph{forward in time} alongside the forward pass. This has been called \emph{Real-Time Recurrent Learning} (RTRL) \citep{williams1989learning}. This property of RTRL is undeniably attractive however it comes with immense computational downsides. While the memory cost of RTRL is fixed (unlike BPTT), it scales cubically with the network dimension which means it only outperforms BPTT at extremely long sequences for most network sizes. Moreover, its' computational cost scales \emph{quartically} with hidden dimension which makes it impractical for all but small networks. This is because RTRL explicitly materializes and operates over the `sensitivity' matrix $\frac{d h}{d\theta}$ where $h$ is the recurrent states and $\theta$ is the full set of network parameters. This matrix measures effectively how every parameter interacts with every hidden state through time. 

To reduce the computational burden of RTRL various approximations have been proposed which usually involve performing some kind of low-rank factorization or approximation of the sensitivity matrix \citep{tallec2017unbiased,benzing2019optimal,menick2020practical,mujika2018approximating}. One such approach is E-prop \citep{bellec2020solution}, which effectively replaces the total derivative in the sensitivity matrix with the partial derivative $\frac{dh}{d\theta} \approx \frac{\partial h}{\partial \theta}$. What this means is that E-prop ignores indirect interactions between parameters mediated by the states across time, which full RTRL correctly tracks. This approximation, although fairly stringent, brings substantial compute and memory benefits. The memory cost of E-prop is simply quadratic in the hidden dimension and hence equivalent to backprop while also being constant in the sequence length, potentially allowing sequences of indefinite length. Computationally, the cost of E-prop matches BPTT while being able to run online in a purely forward fashion. Biologically E-prop is implemented by storing a scalar 'eligibility trace' at every synapse with a local(ish) learning-rule which governs the update of the eligibility trace.

However, E-prop was only ever defined for a single layer of recurrent neurons, and effectively no literature has considered how E-prop could be applied to train deep networks of recurrent neurons where each layer has its own recurrent dynamics. Indeed, naively, it is unclear if this is possible since the interactions engendered between different layers and across time could render the recursive dynamic-programming-style recursion underlying E-prop invalid. \citet{irie2023exploring} who studied the feasibility and effectiveness of RTRL in more modern networks briefly discussed deep networks and came to a similar conclusion.

However, in this mathematical note we demonstrate a straightforward recursive algorithm for performing E-prop credit assignment in a local and online fashion in arbitrarily deep networks and indeed across arbitrary DAG computation graphs. Our algorithm requires computing and storing an eligibility trace at each synapse for every hierarchical layer.

\section{Preliminaries}

\subsection{BPTT}

\begin{figure}
    \centering
    \includegraphics[width=0.5\linewidth]{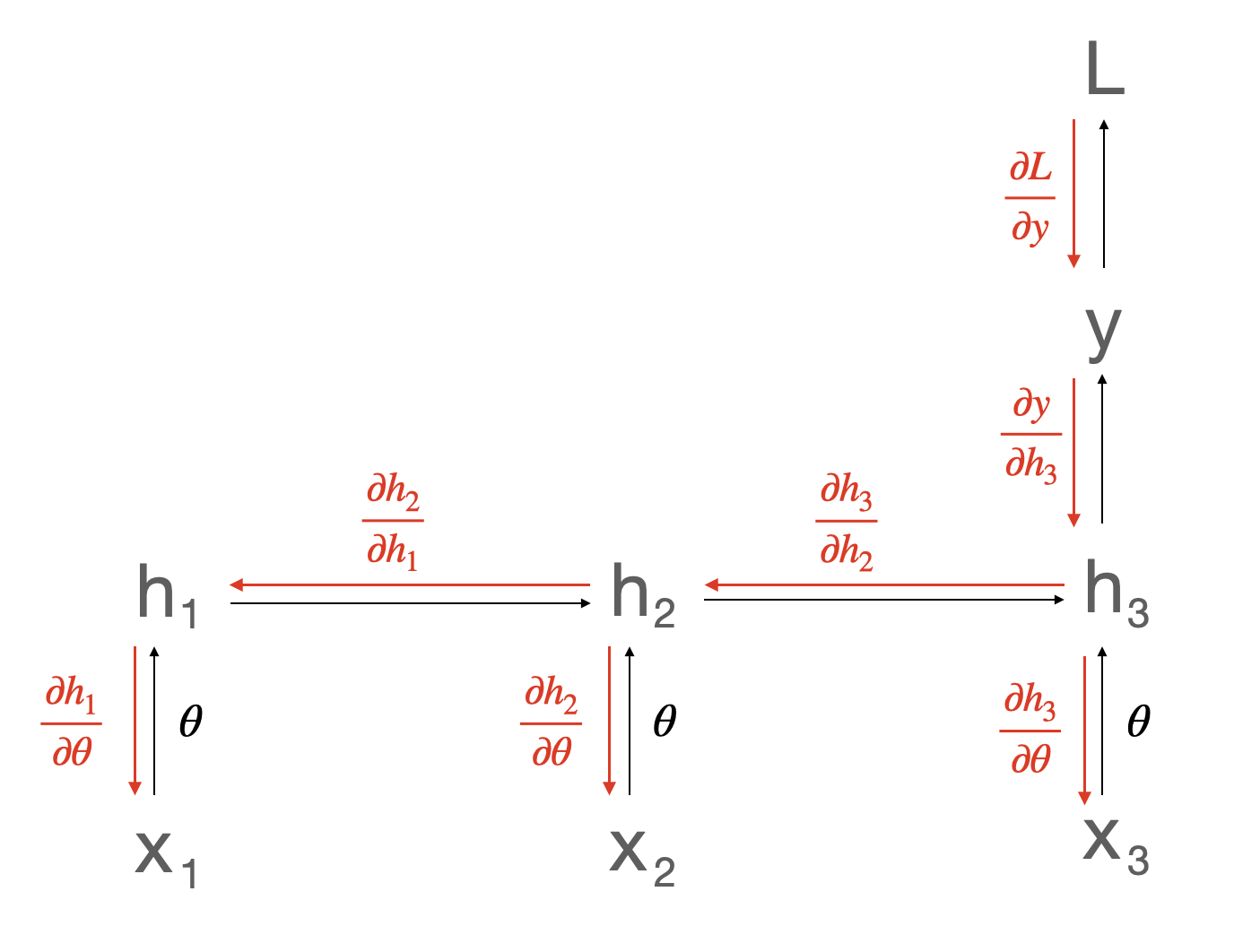}
    \caption{Schematic computation graph of a single layer RNN model of the type generally used in the RTRL literature. The BPTT gradients are marked in red.}
    \label{fig1}
\end{figure}
Let us consider a very simple RNN operating over three separate timesteps. The computational graph for this model is depicted in Figure \ref{fig1}. Specifically, we have inputs $x_t$ entering the model at each timestep which are mapped to hidden state $h_t$ through some set of parameters $\theta$. The hidden states also depend on previous hidden state $h_t = f(h_{t-1})$ through some recurrent dynamics function $f(\cdot)$. The hidden states are then mapped to output states $y$ which are used to compute the loss $L$. To keep things simple we assume that the loss is only taken at the final timestep, although our results generalize to the case where there is a loss at any set of timesteps up to and including all timesteps. 

In BPTT, we first perform the full forward pass of the model as it iterates through time until it reaches the final iteration. At that point, we stop and compute the loss and then begin the backward phase. The backward phase involves effectively performing the forward pass `in reverse' as we step back through the network going backwards in time computing the gradients of each operation sequentially backwards from the output. 

For instance, BPTT would first compute the gradient from the final timestep as
\begin{align}
    \frac{dL}{d\theta_3} = \frac{dL}{dy_3}\frac{dy_3}{dh_3}\frac{dh_3}{d\theta}
\end{align}
Then it would step back through the recurrent step to timestep two as,
\begin{align}
    \frac{dL}{d\theta_2} = \frac{dL}{dy_3}\frac{dy_3}{dh_3}\frac{dh_3}{dh_2}\frac{dh_2}{d\theta}
\end{align}
And then similarly backwards again to the first timestep. This algorithm thus requires running the network `in reverse' through time to obtain all the required gradients in sequence and also requires storing all the activations for every timestep in memory so they can be used to compute the required derivatives. This mechanism makes BPTT very challenging, if not impossible, for a computational system like the brain to implement.

\subsection{RTRL}

Let us consider the three backward steps of BPTT focusing only on the parts involving the recurrent dynamics,
\begin{align}
    \frac{dL}{d\theta} = \frac{dL}{dy_3}\frac{dy_3}{dh_3}&\Big[ \frac{dh_3}{d\theta_3} + \frac{dh_3}{dh_2}\frac{dh_2}{d\theta} + \frac{dh_3}{dh_2}\frac{dh_2}{dh_1}\frac{dh_1}{d\theta} \Big]
\end{align}
The primary pattern is simply the addition of an extra $\frac{dh_t}{dh_{t-1}}$ term for every step backward in time. What if instead we tried to compute these gradients in an online, `forward' fashion? To do this, we define an recursive trace term $\epsilon_t$. To make this term compute the correct gradients we use the following recursion,
\begin{align*}
    \epsilon_t &= \frac{dh_t}{d\theta} + \frac{dh_t}{dh_{t-1}}\epsilon_{t-1} \\
    \epsilon_0 &= \frac{dh_0}{d\theta} \numberthis
\end{align*}

Notice that our $\epsilon_t$ traces can be computed in an \emph{online} fashion alongside the forward pass since they require no information about the loss. They simply accumulate the sum of inter-timestep gradients about activations which are always available locally at each timestep.

To compute the correct loss gradient we then simply have to multiply the eligibilty trace with the gradient to the eligibility trace from the loss, which is not recurrent,
\begin{align}
    \frac{dL}{d\theta} = \frac{dL}{dy_T}\frac{dy_T}{dh_T}\epsilon_T 
\end{align}
where $T$ denotes the final timestep. This is the RTRL algorithm.

The challenge of RTRL is the `sensitivity' tensor $\frac{dh}{d\theta}$, which where $\theta$ is assumed to be a weight matrix of size $H\times H$ is an $H^3$ tensor since it quantifies the gradient of \emph{every} hidden state against \emph{every} weight. Since the sensitivity matrix is of this dimension the eligibility trace $\epsilon$ is also of the same dimensionality. The key computational operation in RTRL is the computation of $\frac{dh_t}{dh_{t-1}}\epsilon_{t-1}$ which involves the multiplication of an $H \times H$ matrix (the recurrent dynamics matrix $\frac{dh_t}{dh_{t-1}}$ with an $H \times H \times H$ tensor of the previous eligibility trace. This matrix multiplication is $\mathcal{O}(H^4)$ and is the dominant term in the complexity of RTRL. Quartic complexity in the network dimension effectively renders full RTRL infeasible for large networks

\subsection{E-prop}

E-prop simply applies the standard RTRL recursion with one modification: replacing the total derivative with the partial $\frac{dh}{d\theta} \approx \frac{\partial h}{\partial \theta}$. The E-prop iteration thus becomes,
\begin{align}
        \epsilon_t &= \frac{\partial h_t}{\partial \theta} + \frac{\partial h_t}{\partial h_{t-1}}\epsilon_{t-1}
\end{align}
While this distinction seems subtle, it is actually of immense importance. What this means in practice is the E-prop ignores all indirect influences of one parameter on another mediated by the states across time. 

To make this concrete, let's consider the classic RNN case where $\theta$ is a weight matrix connecting inputs $x_t$ to hidden state $h_t$. Here, only a single column of $\theta$ connect to any given hidden state, i.e., $h_{t,i} = \sum_j \theta_{i,j} x_{t,j}$. However, despite this other columns of the weight matrix can influence $h_i$ over time. This is because column $\theta_j$ affects hidden state $h_j$ which then affect $h_i$ at the next timestep through the recurrent dynamics $h_t = f(h_{t-1})$. E-prop ignores these indirect influences entirely -- i.e. it assumes that $\frac{dh_i}{d\theta_{[j,:]}} = 0,  \forall j \neq i$. 

By making this approximation, however, E-prop achieves a substantial reduction in memory and computational cost. The sensitivity matrix drops from shape $H \times H \times H$ to shape $H \times H$, which is of the same dimensionality as the parameters. Thus E-prop uses no more compute than BPTT and requires significantly less space since it updates its eligibility trace online over time (and is thus constant in sequence length) while BPTT requires storing the entire history of activations. 

However, E-prop has never been extended to networks deeper than a single layer of recurrent neurons. It is to this question that we now turn.

\section{Extension to Deep Networks}

\begin{figure}
    \centering
    \includegraphics[width=0.5\linewidth]{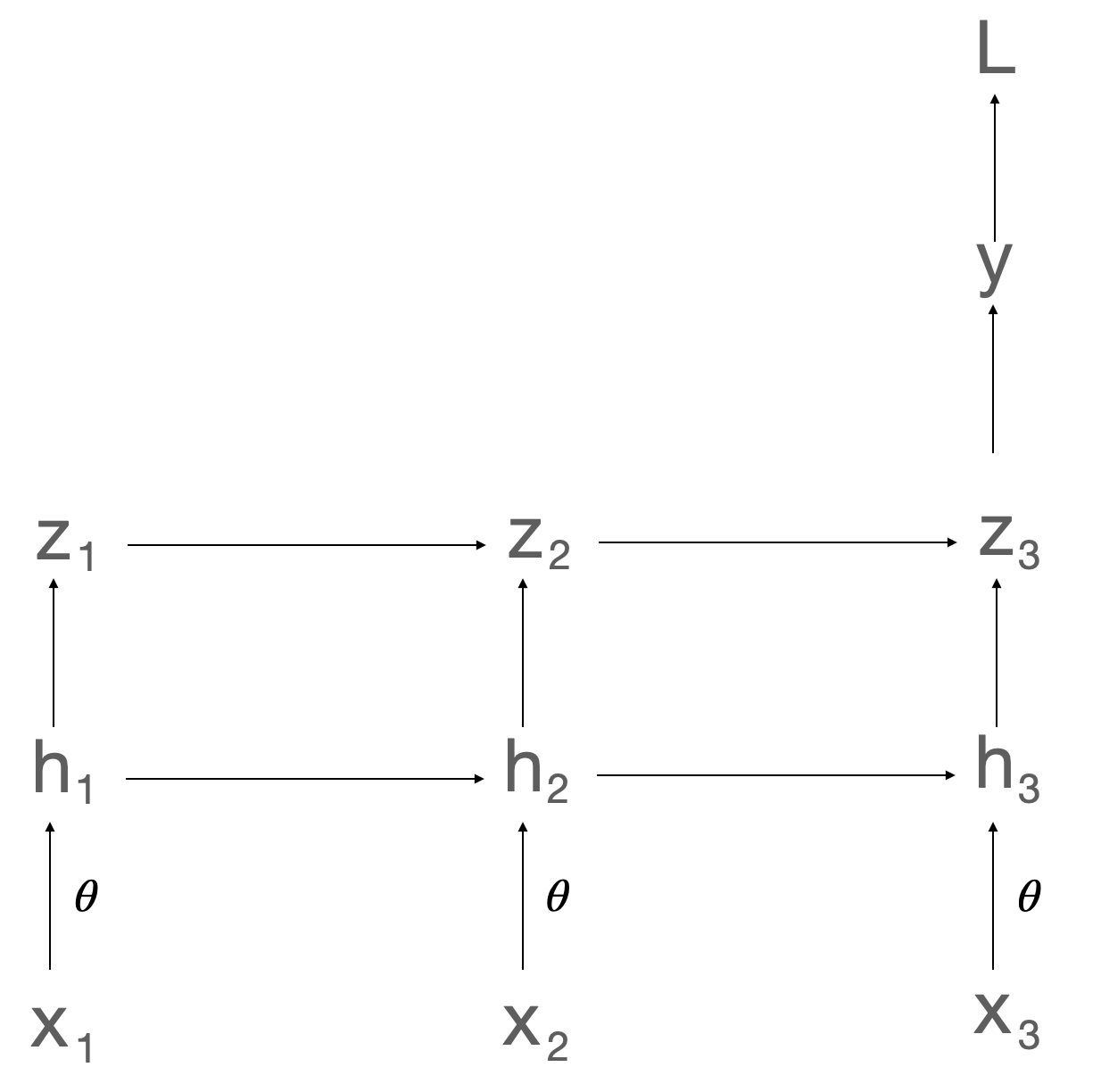}
    \caption{Schematic computation graph of our example two-layer 3-step RNN for which we derive the deep E-prop recursion.}
    \label{fig2}
\end{figure}

Let us start by considering a two-layer network with both layers comprising recurrent units. A schematic of this network is depicted in Figure \ref{fig2}. To derive the recursion, let's start by writing out the full set of gradients for our 2-layer, 3-step network,
\begin{align*}
\frac{\partial L}{\partial \theta}
&= \frac{\partial L}{\partial y}\frac{\partial y}{\partial z_3}\left[
\begin{aligned}
  &\frac{\partial z_3}{\partial h_3}\Big[\frac{\partial h_3}{\partial \theta} + \frac{\partial h_3}{\partial h_2}\big[\frac{\partial h_2}{\partial \theta} + \frac{\partial h_2}{\partial h_1}\frac{\partial h_1}{\partial \theta} \big] \Big] \\
  &+ \frac{\partial z_3}{\partial z_2}\frac{\partial z_2}{\partial h_2}\Big[\frac{\partial h_2}{\partial \theta} + \frac{\partial h_2}{\partial h_1}\frac{\partial h_1}{\partial \theta} \Big] \\
  &+ \frac{\partial z_3}{\partial z_2}\frac{\partial z_2}{\partial z_1}\frac{\partial z_1}{\partial h_1}\frac{\partial h_1}{\partial \theta}
\end{aligned}
\right] \\
&= \frac{\partial L}{\partial y}\frac{\partial y}{\partial z_3}\left[
\begin{aligned}
  &\frac{\partial z_3}{\partial h_3}\epsilon^h_3 \\
  &+ \frac{\partial z_3}{\partial z_2}\frac{\partial z_2}{\partial h_2}\epsilon^h_2 \\
  &+ \frac{\partial z_3}{\partial z_2}\frac{\partial z_2}{\partial z_1}\frac{\partial z_1}{\partial h_1}\epsilon^h_1
\end{aligned}
\right] \numberthis
\end{align*}

We can note that when written like this we can immediately see multiple RTRL style recursions where $\epsilon^h_t = \frac{\partial h_t}{\partial \theta} + \frac{\partial h_t}{\partial h_{t-1}}\epsilon^h_{t-1}$. Now note that once written like this there is also a clear structure in the $\frac{\partial z_t}{\partial z_{t-1}}$ terms. 
\begin{align}
    \frac{\partial L}{\partial \theta}
&= \frac{\partial L}{\partial y}\frac{\partial y}{\partial z_3} \Big[\frac{\partial z_3}{\partial h_3}\epsilon^h_3 + \frac{\partial z_3}{\partial z_2}\big[\frac{\partial z_2}{\partial h_2}\epsilon^h_2 + \frac{\partial z_2}{\partial z_1}\big[\frac{\partial z_1}{\partial h_1}\epsilon^h_1\big]\big]
\end{align}

Which can be re-expressed as a second recursion in the $z$ hierarchical variables. Putting this together, we obtain two nested recursive iterations $\epsilon^h$ which is the trace of the lower layer and $\epsilon^z$ which is the trace for the higher layer,
\begin{align*}
    \epsilon^z_t &= \frac{\partial z_t}{\partial h_t}\epsilon^h_t + \frac{\partial z_t}{\partial z_{t-1}}\epsilon^z_{t-1} \\
    \epsilon_t^h &= \frac{\partial h_t}{\partial \theta} + \frac{\partial h_t}{\partial h_{t-1}}\epsilon^h_{t-1} \numberthis
\end{align*}

One way to think about this is to think about gradient paths through the unrolled network. The unrolled network forms a two dimensional lattice where we can either move horizontally `in time' or vertically `in depth'. We want to walk backwards from the top right (the output of the network at the end of time) to the bottom left (the input of the network at the beginning of time). At every point there are two directions we can move -- either leftwards, backwards in time, or downwards, to a lower layer. Once we have moved, exactly the same choices present themselves to us, until we hit a terminal node at the lowest layer of the network. Turning this around to think of the forward recursion, this means that each node in the lattice must receive two components, one `recurrent' component coming forward in time from the same layer, and one `hierarchical` component coming from a lower layer to a higher one. Since in backprop we simply sum across all gradient paths, all we have to do at each node in the lattice is sum these components multiplied by their derivative of transiting from their node to our node either across time or across depth. We thus arrive at exactly the trace update rule derived above. 

This reasoning also makes it clear that these recurrences apply to general networks of arbitrary depth. To demonstrate this more clearly, we can combine the two recurrences into a single recurrence with a form that depends upon whether we are in a terminal node or not. 
\begin{align*}
    \epsilon_t^l &= \frac{\partial h^l_t}{\partial h^l_{t-1}} \epsilon^l_{t-1} + K \\
    K &= \frac{\partial h^l_t}{\partial h^{l-1}_t} \epsilon^{l-1}_t \, \, \, \text{if nonterminal} \\
    K &= \frac{\partial h^l_t}{\partial \theta} \, \, \, \text{if terminal} \numberthis
\end{align*}

where $h^l_t$ denotes the hidden recurrent states of layer $l$ at timestep $t$. This learning rule requires only a single trace per parameter group and is both very simple and completely local assuming that each synapse has access to the derivatives of the recurrent and hierarchical dynamics.  

To generalize this result to the case where we have a loss computed on the output of every step we simply sum the gradients of the output losses at every timestep computed against the final layer traces at that timestep. 

\subsection{Extension to arbitrary DAGs}

This general way of thinking about gradient paths gives us an immediate generalization to the case of arbitrary DAG networks where each node in the DAG also has internal recurrence. At every node, we have the option of moving to every connecting child node and also to go backwards in time. Thus to construct the recurrence we simply have to sum the traces of all children of our parent node plus its recurrent contribution. 

\begin{align}
    \epsilon_t &= \frac{\partial h^l_t}{\partial h^l_{t-1}} \epsilon_{t-1} + \sum_{i \in \text{child($\epsilon_t$)}} \frac{\partial h_t}{\partial h^i_t} \epsilon_t^i
\end{align}

\section{Discussion}

Here we have demonstrated mathematically that our algorithm performs correct credit assignment across both time (E-prop) and depth, and that this only requires a second eligibility trace operating on simple local learning rules. Such a mechanism seems promising as a potential method for credit assignment to work in the cortex. While we have derived our algorithm in the context of E-prop, it is worth noting that our approach also generalizes to full RTRL where we simply replace the partials with full derivatives in our iteration expressions.

Importantly, despite prior concerns \citep{irie2023exploring} about the multi-layer case, our algorithm, due to the strict approximations inherent in E-prop maintains linear complexity in depth, despite the combinatorial explosion of possible gradient paths that might be naively assumed. This is because our iteration recursively sums across all possible paths entering a node, thus applying a dynamic-programming like strategy to maintain low computational complexity. 

However, the work we have done here is still very preliminary. Firstly, we have performed no experiments demonstrating that good credit assignment across depth works in practice. While we are confident that the full RTRL variant of our algorithm would work, since it is mathematically equivalent to BPTT across both time and depth, it remains unclear how effective the E-prop approximation remains when performing credit assignment across both depth and time for large networks in challenging tasks. While the E-prop paper shows good performance approaching BPTT, this is only on single-layer networks and relatively straightforward tasks. Scaling and testing the performance of approximate algorithms like this on realistic and challenging tasks, especially based on language and requiring both reasoning and long-context recall is an open question in the field. 

Secondly, beyond the quality of its approximations, our proposed algorithm still has several clear limitations which are also inherent in E-prop but which do not become apparent until deeper networks are considered:

\begin{enumerate}
    \item Performing credit assignment with multiple sets of parameters such as different layers with different parameters or separate recurrent, input, and output weights is challenging. Every set of parameters requires a different set of eligibility traces for their credit propagation and so for deep networks this becomes quickly unmanageable in practice. Even single-layer E-prop has this problem if we consider separate recurrent weights.
    \item All the gradient information gets forwarded 'up and to the right' i.e. to the highest layer of the network closest to the loss and at the last time point. To actually perform the weight update, this information would need to be transmitted back to its originating layers somehow.
    \item E-prop does not actually perform online weight updates it accumulates gradient information online across time but must wait until the end of the episode to actually update the weight. It is possible to update weights online but this results in the eligibility traces becoming subtly off-policy since they would then be accumulating gradient information from stale weights. The brain, however, appears to perform synaptic plasticity online as well as credit accumulation
    \item Our approach, like E-prop and almost all other approaches, assumes that the functions in the computation graph are differentiable, and that the brain has mechanisms of computing their derivatives and transporting that information to the synapses correctly so that it can be used to correctly compute the eligibiltry trace at the synapse.
    \item Our approach is online, so does not suffer from backward locking, however it has the classic weight symmetry issues as standard backpropagation when applied to the usual parametrization of MLPs
\end{enumerate}

\clearpage

\bibliographystyle{tmlr}
\bibliography{main}

\clearpage



\appendices

\end{document}